\documentclass[10pt,journal,compsoc]{IEEEtran}
\pdfoutput=1

\ifCLASSOPTIONcompsoc
  \usepackage[nocompress]{cite}
\else
  \usepackage{cite}
\fi

\ifCLASSINFOpdf
\else
\fi

\usepackage{amsmath}
\usepackage{amsfonts}
\usepackage{booktabs} 
\usepackage[english]{babel}
\usepackage{varioref}
\usepackage{import}
\usepackage{subfiles}
\usepackage{array}
\usepackage{makecell}
\usepackage{booktabs}
\usepackage{multirow}
\usepackage{hhline}
\usepackage{subcaption}
\usepackage{caption}
\usepackage{textcomp}
\usepackage{gensymb}
\usepackage{graphicx}
\usepackage{comment}

\DeclareMathOperator*{\argmax}{argmax}
\DeclareMathOperator{\E}{\mathbb{E}}

\usepackage{array}
\usepackage{url}

\usepackage{etoolbox}

\usepackage{hyperref}
\usepackage{cleveref}

\begin{document}

%
\title{NeuSaver: Neural Adaptive Power Consumption Optimization for Mobile Video Streaming}
%
%
%
\author{Michael~Shell,~\IEEEmembership{Member,~IEEE,}
        John~Doe,~\IEEEmembership{Fellow,~OSA,}
        and~Jane~Doe,~\IEEEmembership{Life~Fellow,~IEEE}
\IEEEcompsocitemizethanks{\IEEEcompsocthanksitem M. Shell was with the Department
of Electrical and Computer Engineering, Georgia Institute of Technology, Atlanta,
GA, 30332.\protect\\
E-mail: see http://www.michaelshell.org/contact.html
\IEEEcompsocthanksitem J. Doe and J. Doe are with Anonymous University.}
\thanks{Manuscript received April 19, 2005; revised August 26, 2015.}}

\author{Kyoungjun~Park,~Myungchul~Kim,~\IEEEmembership{Member,~IEEE,}~and~Laihyuk~Park
\IEEEcompsocitemizethanks{\IEEEcompsocthanksitem K. Park is with the TmaxSoft, Inc., Gyeonggi-do 13595, South Korea. E-mail: \href{mailto:kyoungjun_park@tmax.co.kr}{kyoungjun\_park@tmax.co.kr}
\IEEEcompsocthanksitem M. Kim is with the Department of School of Computing, KAIST, Daejeon, South Korea. E-mail: \href{mailto:mck@kaist.ac.kr}{mck@kaist.ac.kr}
\IEEEcompsocthanksitem L. Park is with the Department of Computer Science and Engineering, Seoul National University of Science and Technology, Seoul, South Korea. E-mail: \href{mailto:lhpark@seoultech.ac.kr}{lhpark@seoultech.ac.kr}
}
}


%
%

\markboth{IEEE Transactions on Mobile Computing}%
{Submitted paper}
%



\IEEEtitleabstractindextext{%
\begin{abstract}
Video streaming services strive to support high-quality videos at higher resolutions and frame rates to improve the quality of experience (QoE). However, high-quality videos consume considerable amounts of energy on mobile devices. This paper proposes NeuSaver, which reduces the power consumption of mobile devices when streaming videos by applying an adaptive frame rate to each video chunk without compromising user experience. NeuSaver generates an optimal policy that determines the appropriate frame rate for each video chunk using reinforcement learning (RL). The RL model automatically learns the policy that maximizes the QoE goals based on previous observations. NeuSaver also uses an asynchronous advantage actor-critic algorithm to reinforce the RL model quickly and robustly. Streaming servers that support NeuSaver preprocesses videos into segments with various frame rates, which is similar to the process of creating videos with multiple bit rates in dynamic adaptive streaming over HTTP. NeuSaver utilizes the commonly used H.264 video codec. We evaluated NeuSaver in various experiments and a user study through four video categories along with the state-of-the-art model. Our experiments showed that NeuSaver effectively reduces the power consumption of mobile devices when streaming video by an average of 16.14\% and up to 23.12\% while achieving high QoE.
\end{abstract}

}

\maketitle

\IEEEdisplaynontitleabstractindextext
%
\IEEEpeerreviewmaketitle


\IEEEraisesectionheading{\section{Introduction} \label{sec:introduction}}


\IEEEPARstart{V}{ideo} streaming services on mobile devices have witnessed explosive growth in recent years. In 2017, Cisco \cite{CiscoWhitePaper} predicted that mobile video traffic would increase by 55\% annually, comprising 79\% of all mobile traffic by 2022. The main reason for the surge in video traffic is that video streaming services and mobile devices are trying to support higher frame rates and resolutions to improve the quality of experience (QoE) \cite{YouTubeRecommend, SamsungSmartphones}. In addition, human interactive services such as virtual reality and $360\degree$ video streaming, which require high-quality content, are considered as popular applications in mobile infrastructure \cite{lai2019furion, aggarwal2020evaluate, qian2018flare, abari2017enabling}. However, high-quality videos require large amounts of device resources, mainly display-related components, resulting in much higher energy consumption \cite{lim2016adaptive,naderiparizi2018towards}.

Many smartphone manufacturers have endeavored to extend battery lifespans. For example, the battery capacities of Samsung Galaxy smartphones have been increasing at a compound annual growth rate (CAGR) of 7.93\% over the past five years, indicating a steady improvement in battery capacity \cite{SamsungSmartphones}. However, since the volume of mobile video traffic is expected to surge by 55\% annually according to Cisco, the batteries of mobile devices will have difficulty in meeting the high energy consumption requirements of video streaming services in the near future. Furthermore, in the case of small devices such as smartphones and internet of things (IoT) devices, it is impossible to increase the battery capacity to an ideal level because the size of the hardware on which the battery is mounted is quite limited.

To reduce the power consumed by video streaming, it should be noted that not all parts of a video necessarily need high frame rates. In other words, an adaptive frame rate can be applied to each part of the video depending on the degree of motion. For example, a scene in which a tennis player prepares to serve a tennis ball can be applied at a lower frame rate because the degree of change between frames is very small. On the other hand, a higher frame rate should be applied when there is high variation in the scene, such as during the swing motion or the motion of the tennis ball to avoid QoE degradation.


In several works, the frame rate has been dynamically scaled based on the rendered frame contents on the client-side to optimize the power consumption by reducing the operation of display resources on mobile devices \cite{kim2016content, RAVEN, lpgl2019}. The authors measured the similarity between frames and then dropped duplicate or very similar frames. However, these approaches are difficult to apply to video streaming services because they do not consider the unique environment of video streaming such as bit rate adaptation and content delivery network (CDN). To compensate for these shortcomings, another approach was proposed, with a system that effectively applies adaptive frame rates to videos in streaming environments \cite{park2019evso}. This approach eliminates similar video frames on the streaming server to optimize the power consumption of mobile devices without significantly impacting the user experience. However, this system makes improper decisions for some scenes because it determines the frame rate based on a simple heuristic model. Moreover, this system has several practical issues, such as increasing the total number of intra-coded frames by performing video chunking without considering the group of pictures (GOP) structure representing the order of intra- and inter-frames.

In this paper, we propose the NeuSaver system, which determines the proper frame rate for each video chunk and applies adaptive frame rates to videos without user effort or considerable overhead. NeuSaver uses reinforcement learning (RL) techniques \cite{mnih2016asynchronous, mao2017neural, shen2019deepapp, joseph2019towards} to learn an optimal policy \textit{automatically} and \textit{effectively} that determines the appropriate frame rates for each video chunk through a previous learning experience.

The RL model of NeuSaver begins learning without knowing anything about the task and feeds various raw observations such as frame similarities and video chunk sizes to the deep neural network (DNN). The RL model tracks various video characteristics to predict the current situation and to maximize the total reward. In particular, the RL model made more effective decisions than fixed rule-based approaches in various scenes, such as some object movement or camera movement.
NeuSaver trains the DNN using the asynchronous advantage actor-critic (A3C) \cite{mnih2016asynchronous} algorithm, which can quickly reinforce the model by using multiple agents. The NeuSaver simulator allows the RL model to learn about 287 hours of experience in just one minute.

To analyze the motion intensities of video chunks, NeuSaver utilizes the similarity information of video frames generated during the encoding process. When a video is uploaded to the streaming server, the video encoder evaluates the frame similarities based on \textit{macroblocks} in the course of the video compression (or video coding) process. NeuSaver extracts these macroblocks and feeds them to the RL model as the motion intensity information of the video chunks.

We implemented the RL model using TensorFlow \cite{abadi2016tensorflow} and conducted a broad range of experiments with 571 videos. Our experimental results showed that NeuSaver can significantly reduce energy consumption with little effect on the user experience. NeuSaver also outperformed the state-of-the-art model \cite{park2019evso}, which saved an average of 12\% energy, reducing the energy consumption rate by an average of 16.14\% and up to 23.12\%. In particular, NeuSaver saved 8\% of additional energy in the \textbf{News} category compared to the previous study. In addition, the user study showed that users could not clearly distinguish between the original videos and videos processed by NeuSaver. To the best of our knowledge, NeuSaver is the first system to apply the RL approach to achieve energy savings by scaling the frame rate of video chunks based on various video features. The main contributions of this paper are as follows:
\begin{itemize}

  \item We propose a novel deep RL model that selects appropriate frame rates of video chunks based on previous observations.

  \item Similar to the dynamic adaptive streaming over HTTP (DASH, or MPEG-DASH), we present a new adaptive frame rate streaming system that allows the client to select the appropriate frame rate for each video chunk.


  \item We present the design and prototype of a video streaming system utilizing an open H.264 codec to reduce the energy consumption of mobile devices when streaming videos.

  \item We conduct various experiments and a user study showing that NeuSaver significantly reduces the power consumption of mobile devices while preserving QoE.

\end{itemize}




\section{Motivation and Background} \label{sec:motivation}

This section highlights the necessity of NeuSaver and explains the frame characteristics leveraged by this system.

\subsection{Similarity Trend Analysis in Videos}
\begin{figure}[!t]
    \begin{subfigure} {1\columnwidth}
        \includegraphics[width=1\textwidth]{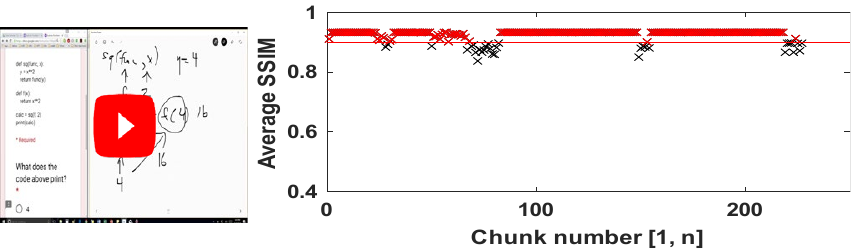}
        \caption{Lecture video (static type).}
        \label{fig:lectureVideo}
    \end{subfigure}
    \begin{subfigure} {1\columnwidth}
        \includegraphics[width=1\textwidth]{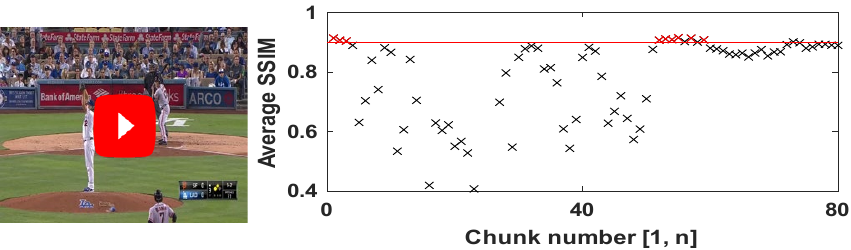}
        \caption{Sports video (dynamic type).}
        \label{fig:sportVideo}
    \end{subfigure}
    \begin{subfigure} {1\columnwidth}
        \includegraphics[width=1\textwidth]{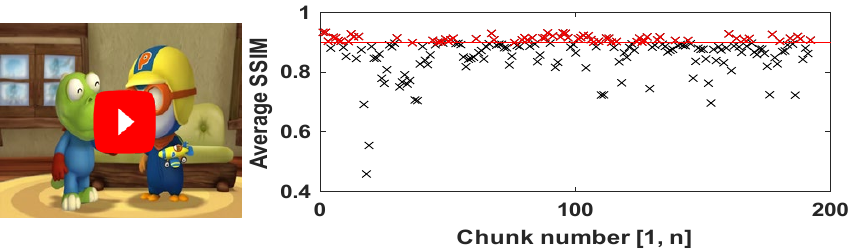}
        \caption{Animated video (hybrid type).}
        \label{fig:animatedVideo}
    \end{subfigure}
    
    \caption{Variation of average SSIM between video chunks for three video types. The length of each video chunk is 2 seconds.}
    \label{fig:SSIMVideos}
\end{figure}

The structural similarity (SSIM) index \cite{SSIM} is mainly used to measure frame similarity. Figure~\ref{fig:SSIMVideos} shows average SSIM values between adjacent frames in video chunks of three different video types. These videos are classified as static, dynamic, and hybrid based on motion intensity. The SSIM index ranges from 0 to 1, and values above 0.9 are considered to indicate strong similarity between frames \cite{Kahawai}. The red shaded marks in Figure~\ref{fig:SSIMVideos} indicate that the average SSIM is greater than 0.9, which means that the degree of change in the video chunk is fairly low. According to Figures~\ref{fig:SSIMVideos}(a) and (b), the average frame similarity varies considerably according to the video type. In most cases, a video contains various motion intensities, as shown in Figures~\ref{fig:SSIMVideos}(b) and (c). For example, in the sports video shown in Figure~\ref{fig:SSIMVideos}(b), video chunks 1-4 have static movements in which the pitcher prepares to throw the ball, while video chunks 5-20 have dynamic movements, where the catcher quickly throws the ball to first base to catch the runner. The similarity in these two scenarios shows a huge variety of more than 60\%, indicating that even within a single video, the degree of change can vary significantly depending on which part of the video is playing. Thus, by lowering the frame rates of the chunks with slow motion, it is possible to reduce the energy consumption of mobile devices while maintaining high QoE.

\subsection{Perceptual Similarity Method}
\begin{figure}[!t]
\centering
    \includegraphics[width=1\columnwidth]{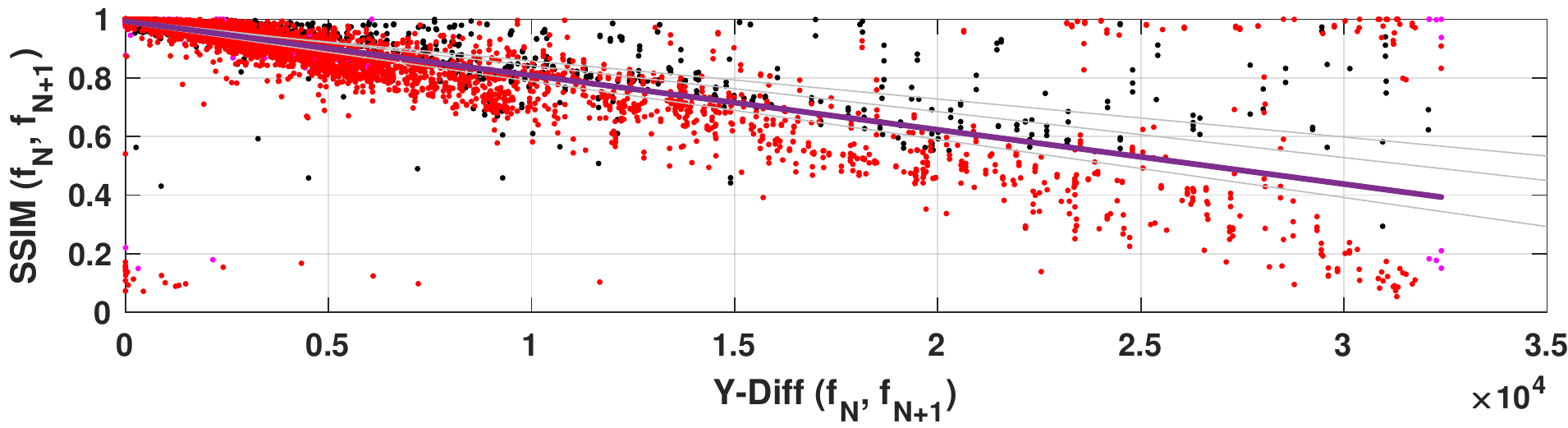}
    \caption{Correlation between Y-Diff and SSIM of three different types of videos (Pearson correlation coefficient: -0.8162).}
    \label{fig:ssim}
\end{figure}

It is important to measure the motion intensities of video chunks accurately and efficiently. The common means of measuring similarity between frames is the SSIM method. However, SSIM involves a high computational cost. It took about 305 milliseconds to measure the similarity between two frames at 1080p resolution using SSIM on a PC with a 3.20 GHz\(\times\)12 processors. Therefore, it took about 3 hours to analyze a single video with a 60 FPS of 10 minutes using SSIM. It is impractical to use SSIM to calculate the similarity of every frame in a video streaming server, where hundreds of hours of video are uploaded every minute \cite{chang2019lsim, RAVEN, park2019evso}.

Park et al. \cite{park2019evso} extracted the luminance values when an H.264/AVC encoder performed video compression to transcode the initial video into the desired one on the streaming server. The H.264/AVC processed each coded picture through a block form called a \textit{macroblock}.
The macroblock consists of luminance (Y) samples and chrominance (Cb, Cr) samples in the YCbCr color space. The authors focused on the fact that the Y samples had more weight than the Cb and Cr samples since the human visual system is much more sensitive to luminance than it is to chrominance \cite{Vision}. NeuSaver extracts and utilizes these Y samples to analyze the motion intensities of video chunks efficiently and accurately. The additional time overhead it took to extract the luminance values between macroblocks of a single video is only 170 milliseconds, which is much less than the SSIM calculation. In addition, the Y-Diff metric that determines the similarity between frames through the Y value difference is quite similar to SSIM. Figure~\ref{fig:ssim} shows the Pearson coefficient correlation (PCC) between SSIM and Y-Diff of the three videos used in Figure~\ref{fig:SSIMVideos}. PCC between SSIM and Y-Diff is about 81\%, showing a meaningful correlation. Note that the video codec of NeuSaver is not required to be H.264/AVC, as other video codecs use the Y values of macroblocks to perform the video compression process.


\section{Overview} \label{sec:overview}

\begin{figure}[!t]
\centering
    \includegraphics[width=1\columnwidth]{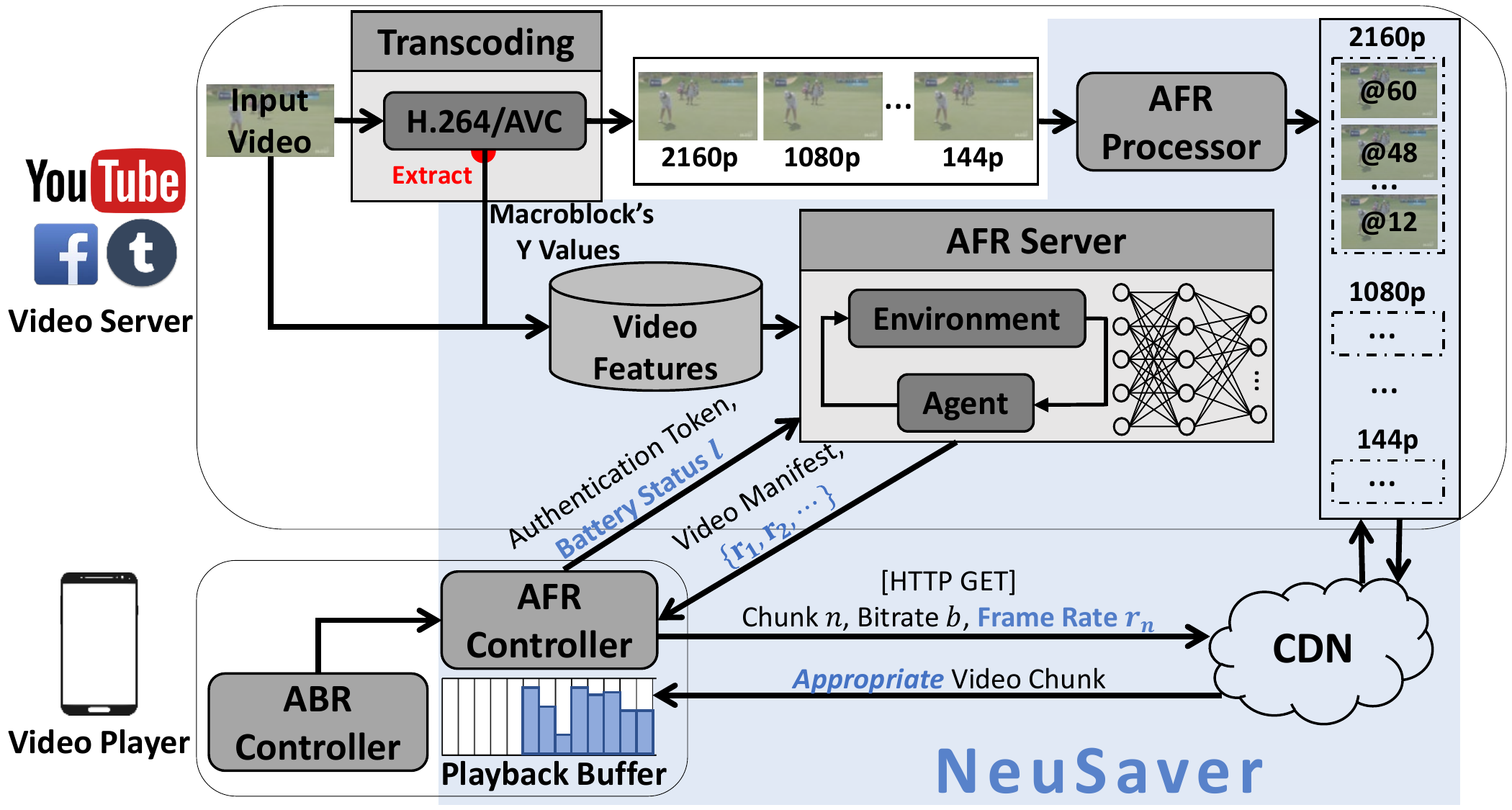}
    \caption{Overview of adaptive video streaming with NeuSaver.}
    \label{fig:NeuSaverArchitecture}
\end{figure}
DASH \cite{DASH}, also called MPEG-DASH, enables adaptive streaming services considering various situations of mobile devices such as network throughput and buffer status. To support this feature, the video server should \textit{transcode} the uploaded video into the desired video set, such as 1080p, 144p, and deliver it to the assigned content delivery network (CDN). Depending on the design of the streaming system, the transcoding process can also be performed on the CDN to distribute the workload of the video server. This series of steps enables the user to request the bit rate of the video chunk suitable for the current situation through the \textit{adaptive bit rate (ABR) controller} \cite{yin2015control, spiteri2016bola, mao2017neural, akhtar2018oboe, yan2020learning}.

Figure~\ref{fig:NeuSaverArchitecture} illustrates the operation of NeuSaver in the traditional video streaming process. The light blue shaded areas represent additional modules and procedures required to support NeuSaver. NeuSaver consists of three major components: \textit{adaptive frame rate (AFR) controller}, \textit{AFR server}, and \textit{AFR processor}. When the server transcodes the uploaded video to create videos with various types of resolution for DASH functionality, NeuSaver extracts the Y values of macroblocks used for video transcoding and uses it to measure perceptual similarities of video chunks. Using the extracted Y values and other video features, the AFR server trains the RL model that determines the appropriate frame rate for video chunks. In addition, the AFR processor additionally performs the processing of generating video chunks at multiple frame rates for a video transcoded at different resolutions. We set up to generate five frame rate levels in our experiment. For example, the AFR processor additionally generates videos with frame rates of 12, 24, 36, and 48 for each resolution for video with a 60 frame rate. The implementation details of the AFR processor are provided in Section~\ref{sec:implementation}.

The AFR server consists of two RL models intended for different purposes: quality assurance and energy reduction. When the AFR controller sends a token to the video server for authentication, it also encloses the battery status \(l\) of the mobile device. The AFR server operates an appropriate RL model that matches the battery status \(l\) based on a preset threshold. The AFR server then responds to the AFR controller with the appropriate set of frame rates and a video manifest containing the location of the CDN hosting the video and a list of available bit rates. Through the above series of steps, users can request the CDN for a video chunk \(n\) that matches the bit rate \(b\) calculated by the ABR controller and the current battery status \(l\) of the video player, as measured by the AFR controller.

The AFR server aims to solve the following trade-off problem:
\begin{equation}
 \label{eq:final_goal}
 \argmax_{r \in F} \quad \mu_{a} N(Q) - \mu_{b} N(E)
\end{equation}
\noindent where \(Q\) and \(E\) represent the perceived video quality and energy cost of the mobile device, respectively; \(\mu_{a}\) and \(\mu_{b}\) are the scaling factors that apply weights according to the importance of the variables; and \(N(x)\) is a min-max normalization function, i.e., \(N(x) = (x - x_{min}) / (x_{max} - x_{min})\), that is used to standardize the characteristics of independent variables to the same scale. This scaling feature also has the advantage that the gradient descent algorithm used in the RL model converges more quickly \cite{ioffe2015batch}. The goal of the AFR server is to select an appropriate frame rate \(r\) from the set of all possible frame rates \(F\) that can reduce energy requirements with little impact on the QoE.

We separated the process of the ABR controller and the AFR controller to facilitate system integration by removing the dependency with the existing ABR controller. The AFR controller independently determines the frame rate after the ABR controller determines the bit rate. The decision of the AFR controller may have some influence in choosing proper bit rates; however, the AFR controller only modifies inter-frames (B- and P-type), which occupy a very small capacity, except for intra-frames (I-type), which store most of the video information. The decision of the AFR controller has little effect on the network bandwidth situation, so it is designed to be performed independently without coupling with the ABR controller. We discuss the effect of the AFR controller's decision on the ABR controller in more detail in Section~\ref{subsec:relationship}. Therefore, the QoE metric mentioned in this paper only refers to the decrease in user experience caused by the decision of the AFR controller, excluding various situations such as video stall and jitter caused by the decision of the ABR controller.


\noindent \textbf{How about applying a mapping heuristic method when determining frame rates:}
For about 50,000 experimental video chunks, the Pearson correlation coefficient between the reduced video quality measured using the video multimethod assessment fusion (VMAF) \cite{VMAF} when the FPS was lowered to 20\% and the Y value was 64.05\%. This means that there are several scenes where the Y value does not fully represent the video quality. This is because the Y value is a value in which detailed information is lost, generated by adding the difference in luminance of macroblocks within two frames. Moreover, it is also very important to consider the surrounding frames as video takes the form of a sequential context. Therefore, we used the RL model that not only considers the luminance values of each macroblock unit but also understands the context before and after the video chunk. As a result, the RL model outperformed the heuristic method in all video categories (see Section~\ref{sec:evaluation}).

\section{RL Model} \label{sec:reinforcement}

This section describes the design of the RL-based AFR algorithm that can solve the trade-off problem in Equation~\ref{eq:final_goal}.

\subsection{Training Methodology} \label{subsec:Training}

\noindent \textbf{Choice of the learning algorithm.}
The objective of RL is to find an optimal policy that maximizes the expected cumulative reward in the problem environment represented by the Markov decision process (MDP), which is \((S,A,P_{a},R_{a},\gamma\)). RL has the advantage that the agent solves the specific problem by interacting with the environment without any a priori knowledge about the problem situation (i.e., MDP elements).

There are various RL algorithms for training the learning agents (e.g., DQN \cite{mnih2015human}, DDPG \cite{lillicrap2015continuous}, and Rainbow \cite{hessel2018rainbow}). In NeuSaver, A3C \cite{mnih2016asynchronous} was adopted to train agents for the following reasons: (1) A3C is a state-of-the-art algorithm and successfully solves many learning problems \cite{mao2017neural, huang2018qarc, pang2019towards, sassatelli2019user}; (2) A3C can quickly converge to the optimal policy because multiple agents are trained concurrently; and (3) A3C can train neural networks (NNs) in a stable way because it uses both value function and policy function to compensate for the weaknesses of the other (e.g., high variance in policy gradient). 
Therefore, we selected A3C as the training algorithm for the RL model to solve the problem in which many users stream numerous kinds of videos from the streaming server.

\noindent \textbf{Construction of the streaming simulator.}
If NeuSaver trains the RL model in a real video streaming environment, the model can only learn as much as the experience it spent for training. Therefore, NeuSaver trains the RL model by constructing a streaming simulator that works similarly to an actual streaming service for efficient learning. For each video chunk download, the NeuSaver simulator takes the Y values and the size of the next video chunk, which were extracted and stored when the video was uploaded to the streaming server. The simulator feeds these state observations (e.g., similarities of the nearby video chunks, sizes of the next chunk, and original frame rate of the video) to the RL agent for training. This chunk-based simulator allows NeuSaver to learn about 287 hours of experience in just one minute.

\subsection{A3C Training Algorithm Basics} \label{subsec:Algorithm}
\begin{figure}[!t]
\centering
    \includegraphics[width=1\columnwidth]{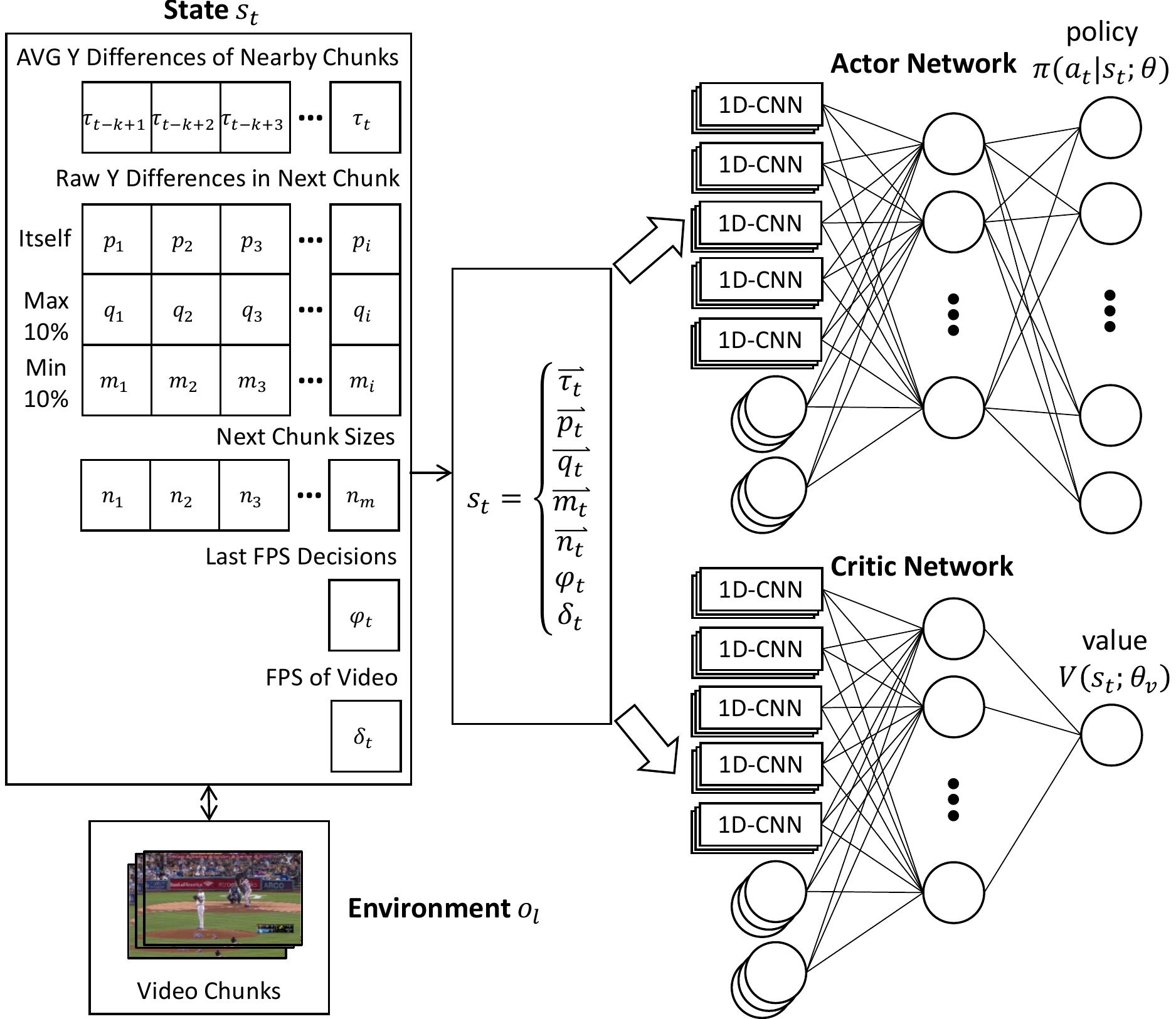}
    \caption{Overall architecture of learning agent that generates the AFR policy based on observations.}
      \label{fig:a3c}
\end{figure}
This section describes the A3C training algorithm, which trains two NNs as shown in Figure~\ref{fig:a3c}.

\noindent \textbf{State.}
After the simulator downloads each video chunk \(t\), the learning agent feeds state inputs \(s_{t} = (\vec{\tau_{t}}, \vec{p_{t}}, \vec{q_{t}}, \vec{m_{t}}, \vec{n_{t}}, \varphi_{t}, \delta_{t})\) provided by the environment \(o_{l}\) to the actor and critic networks. \(o_{l}\) is constructed by randomly selecting videos from the training dataset for each agent; \(\vec\tau_{t}\) is the vector of the average of the Y differences for \(k\) video chunks adjacent to the next video chunk, where \(k\) was set to 2 to account for the similarity of the chunks before and after the next video chunk; \(\vec{p_{t}}\) is the vector of the raw Y differences between all \(i\) frames within the next video chunk, where \(i\) depends on the original frame rate of the video; \(\vec{q_{t}}\) and \(\vec{m_{t}}\) are vectors of the top 10\% and the bottom 10\% of the raw Y difference values between all \(i\) frames within the next video chunk, respectively. \(\vec{n_{t}}\) is the \(m\)-sized vector for the available next video chunks, which \(m\) indicates the degree of variability in the frame rates of the video supported by NeuSaver; \(\varphi_{t}\) and \(\delta_{t}\) represent the last decision of FPS and the original FPS of the video, respectively. Note that we feed the raw Y difference as well as the average Y difference used in the existing approaches. In particular, we added the min and max 10\% values of the raw Y difference so that the RL model can interpret various scenes such as camera movement and movement of some objects. \(\vec{\tau_{t}}\), \(\vec{p_{t}}\), \(\vec{q_{t}}\), \(\vec{m_{t}}\), and \(\vec{n_{t}}\) are passed through the 1D convolutional neural network (CNN) because they are vector types with multiple sequential values. 

\noindent \textbf{Policy and NN structure.}
The learning agent selects the frame rate of the next video chunk suitable for \(s_{t}\) based on the policy \(\pi\). The policy is defined as a probability distribution over actions. \(\pi(a_{t}|s_{t})\) represents the probability of choosing action \(a_{t}\) in state \(s_{t}\). However, state \(s_{t}\), which contains multiple components, has numerous possible cases, resulting in an unmanageable number of \(\{s, a\}\) pairs. Several approaches resolved large-scale problems with significant amounts of \(\{s, a\}\) pairs by applying NNs to the RL model \cite{mao2017neural, liu2018demand, mnih2015human, silver2016mastering}. Therefore, NeuSaver designs the structure of actors and critics as NN to feed raw observation data.

\noindent \textbf{Policy training.}
NeuSaver parametrizes the policy \(\pi\) using \textit{policy parameter} \(\theta\) to train the policy directly. The goal of the RL agent is to update the policy parameter \(\theta\) gradually to find the optimal policy. NeuSaver applies a commonly used \textit{policy gradient} \cite{sutton2000policy} to update \(\theta\). The policy gradient method is used to optimize the parametrized policy for the expected total reward by gradient descent. The gradient of the cumulative reward from the time step \(t\) with discount factor \(\gamma \in (0,1]\) can be calculated as follows:
\begin{equation}
    \nabla_{\theta}\E_{\pi}\big[\sum_{k=0}^{\infty} \gamma^{k}r_{t+k}\big] = \E_{\pi}\big[\nabla_{\theta} \log\pi(a|s;\theta)A(s,a;\theta,\theta_{v})\big]
\end{equation}
\noindent where \(A(s,a;\theta,\theta_{v})\) is the advantage function, which represents the improvement compared to the expected action \(a\) of an agent in a state \(s\). In practice, \(A(s,a;\theta,\theta_{v})\) can be calculated empirically through \(k\) trajectory samples obtained from experience:
\begin{equation}
\label{eq:advantage}
    A(s,a;\theta,\theta_{v}) = \sum_{i=0}^{k-1} \gamma^{i}r_{t+i} + \gamma^{k}V(s_{t+k};\theta_{v}) - V(s_{t};\theta_{v})
\end{equation}
\noindent where \(k\) can vary from state to state depending on how many trajectory samples the agent has when calculating the advantage function, and \(V(s_{t};\theta_{v})\) is a value function estimate. Therefore, each gradient update of the actor-network policy parameter \(\theta\) is
\begin{equation}
\label{eq:policy_paramter}
    d\theta \leftarrow d\theta + \alpha\sum_{t}\nabla_{\theta} \log\pi(a_{t}|s_{t};\theta)A(s_{t},a_{t};\theta,\theta_{v})
\end{equation}
\noindent where the hyperparameter \(\alpha\) is the learning rate that decides to what extent to trust the newly acquired information compared to the old information. \(\nabla_{\theta} \log\pi(a_t|s_t;\theta)\) indicates in which direction the policy parameter \(\theta\) is to be changed to increase the probability of selecting action \(a_{t}\) in state \(s_{t}\). \(A(s_{t},a_{t};\theta,\theta_{v})\) indicates how advantageous it is to step in this direction. In other words, this approach allows the policy to be strengthened where the predictions were lacking.

When calculating the advantage function to update the policy parameter \(\theta\), the agent must estimate the value function, as shown in Equation~\ref{eq:advantage}. The critic network estimates this value function so that the actor network can calculate the advantage function. Similarly to the policy parameter \(\theta\), the value function is parameterized so that the critic network can be updated directly through the \textit{value parameter} \(\theta_{v}\). NeuSaver trains \(\theta_{v}\) using the \textit{temporal difference} \cite{sutton2018reinforcement}. Each update of the value parameter \(\theta_{v}\) of the critic network is
\begin{equation}
\label{eq:value_parameter}
    d\theta_{v} \leftarrow d\theta_{v} + \alpha'\sum_{t} \nabla_{\theta_{v}} \big(r_{t} + \gamma V(s_{t+1};\theta_{v})-V(s_{t};\theta_{v}) \big)^2
\end{equation}
\noindent where the hyperparameter \(\alpha'\) is the learning rate for the critic network.

Finally, adding policy entropy when updating the policy parameter can improve the \textit{exploration} of the agent and reduce the possibility of convergence to a suboptimal solution \cite{mnih2016asynchronous}. Therefore, we modified Equation~\ref{eq:policy_paramter} to include the entropy regularization term as follows:
\begin{multline}
\label{eq:final_policy_parameter}
    d\theta \leftarrow d\theta + \alpha\sum_{t}\nabla_{\theta} \log\pi(a_{t}|s_{t};\theta)A(s_{t},a_{t};\theta,\theta_{v})\\ 
    + \beta\nabla_{\theta}H(\pi(s_{t};\theta)
\end{multline}
\noindent where the hyperparameter \(\beta\) adjusts the intensity of the entropy term and \(H\) is the entropy of the policy.

In summary, the critic network only helps the actor network to update the policy parameter. Note that the actor network determines the proper frame rate in the AFR server. The detailed process of this derivation can be found in \cite{mnih2016asynchronous}.

\noindent \textbf{Parallel training.}
\begin{figure}[!t]
\centering
    \includegraphics[width=1\columnwidth]{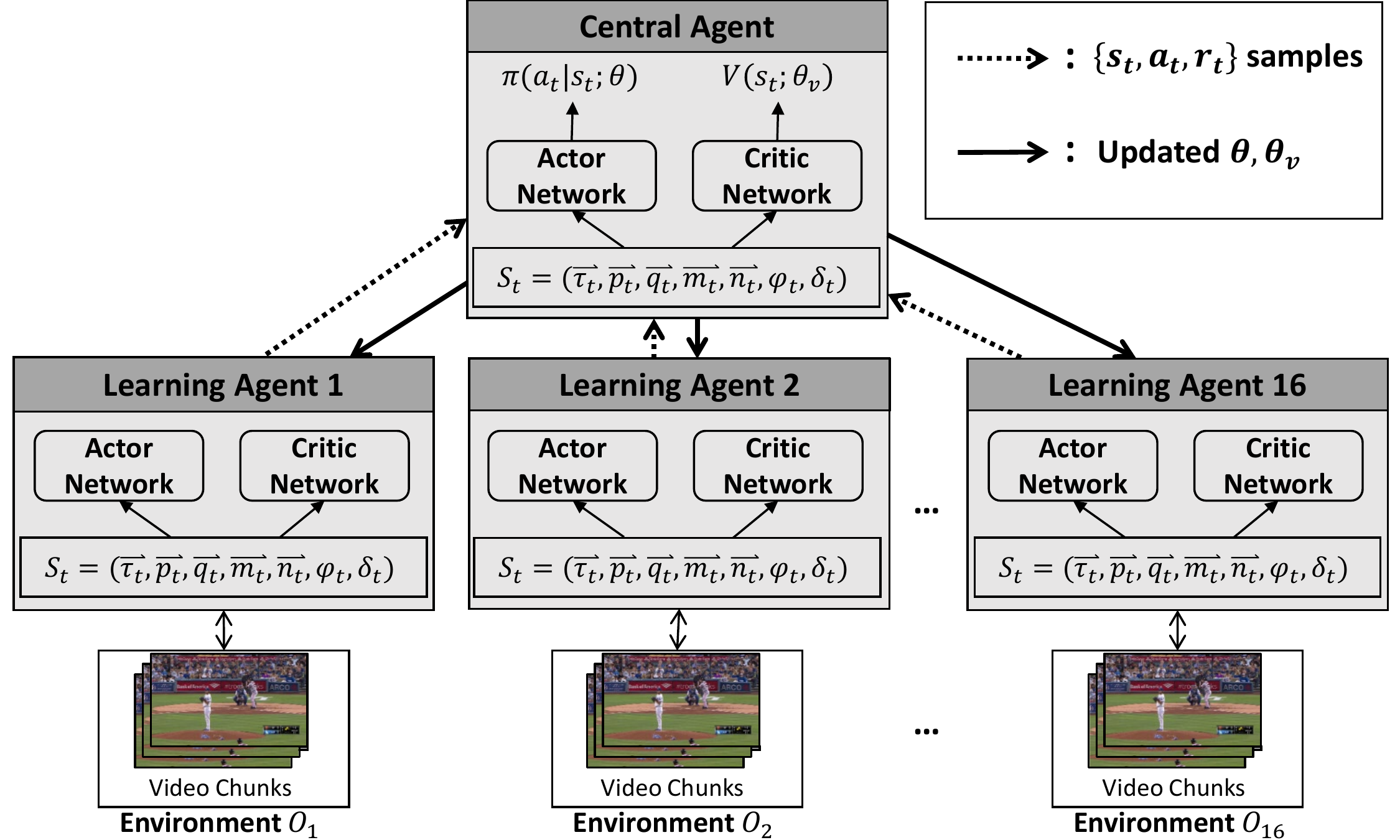}
    \caption{NeuSaver's parallel training design with 1 central agent and 16 learning agents.}
    \label{fig:parallel}
\end{figure}
NeuSaver trains in parallel through 16 learning agents as shown in Figure~\ref{fig:parallel}. The learning agents learn through each differently configured environment and periodically deliver \(\{s_{t}, a_{t}, r_{t}\}\) samples to the central agent. Then, the central agent asynchronously updates the policy parameter \(\theta\) and value parameter \(\theta_{v}\) using the samples received from the learning agents. The parameters are updated using Equations~\ref{eq:value_parameter} and~\ref{eq:final_policy_parameter}. Finally, the central agent delivers the updated parameters to the learning agents, allowing the learning agents to learn using the updated parameters. All of these processes can be performed asynchronously without locking problems between the agents \cite{recht2011hogwild}.

\subsection{Reward Function Design} \label{subsec:Reward}
To solve the optimization problem shown in Equation~\ref{eq:final_goal}, we designed the reward function considering various QoE metrics. The reward function \(R\) for a video with \(N\) chunks is defined as:
\begin{multline}
\label{eq:reward_function}
R = \sum_{k=1}^{N} \big[ \mu_{1}Q(C_{k}^{r}) + \mu_{2}Q_{diff}(C_{k}^{r}) + B(C_{k}^{r}) \\  - P(C_{k}^{r}) - \mu_{3}E_{FPS}(C_{k}^{r}) \big]
\end{multline}
\noindent where \(C_{k}^{r}\) represents video chunk \(k\) with frame rate \(r\) that is selected by the AFR server. \(Q(C_{k}^{r})\) is the video quality of chunk \(C_{k}^{r}\) that is measured using the video multimethod assessment fusion (VMAF) video quality metric \cite{VMAF}. \(Q_{diff}(C_{k}^{r})\) represents the quality difference between chunk \(k\) with the frame rate selected by the AFR server and chunk \(k\) with a frame rate one level lower than \(C_{k}^{r}\), i.e., \(Q_{diff}(C_{k}^{r})=Q(C_{k}^{r})-Q(C_{k}^{r-1})\). If the AFR server picked the lowest frame rate in chunk \(k\), then we set \(Q_{diff}(C_{k}^{r})=0\). The larger \(Q_{diff}(C_{k}^{r})\), the greater the video quality loss when selecting a frame rate lower than that of the AFR server. 
\(B(C_{k}^{r})\) is the bonus score given if the selected video chunk has a high video quality even though the frame rate is lowered.
\(P(C_{k}^{r})\) is the penalty score when the selected video chunk does not meet the target quality. This penalty score serves as the minimum boundary for video quality when the AFR server determines the frame rate. \(E_{FPS}(C_{k}^{r})\) represents the energy consumed to playback chunk \(C_{k}^{r}\), e.g., decode, display. \(E_{FPS}(C_{k}^{r})\) represents the frame rate \(r\) chosen for the video chunk because the energy consumption increases in proportion to the frame rate \cite{kim2016content}. All of the above factors included in the reward function were min-max normalized for effective comparison and model performance as described in Section~\ref{sec:overview}.
\(\mu_{1}\), \(\mu_{2}\), and \(\mu_{3}\) are scaling factors that can balance the importance of each QoE metric. Since the scaling factors determine the characteristics of the learning model, various experiments were conducted, as described in Section~\ref{subsec:Methodology}.

\subsection{Support for Various Situations}
\label{subsec:Support}
The number of neurons in the output layer of actor networks represents the frame rate levels that the RL model can choose. We designed the output layer of the actor network to be five neurons, allowing for five choices, i.e., 12, 24, 36, 48, and 60 for video with a frame rate of 60. However, this design has some practical issues: 1) The original frame rates of the videos can vary, such as 24 or 60, and 2) The range of choices for the desired frame rate can vary depending on the particular scenario. For example, in certain circumstances, a video with a frame rate of 24 can be processed to support only three frame rate selections, such as 8, 16, and 24. A naive means of addressing these issues is to train the models separately for each video property and scenario. This solution may work well with specific video types or in certain situations; however, it is not scalable and effective in general problems.

For a single RL model to meet various requirements, if the range of choices offered by NeuSaver differs from the range required by the target, the selection range can be converted into a target range using the following equation: \(a' = \frac{|T|}{|D|} \times a\). Here, \(a\) is the selection of the RL model and \(a'\) is the selection transformed to fit the target range. \(|T|\) and \(|D|\) are the numbers of possible choices for the target and base models, respectively. \(|D|\) was set to 5 in NeuSaver. Therefore, NeuSaver can adapt to any video type or scenario without further modification.

\section{Implementation} \label{sec:implementation}

We implemented a video streaming server with the AFR server through IIS \cite{IIS}. To operate 16 learning agents in parallel, we configured a server equipped with 3.20 GHz\(\times\)12 processors and 16 GB RAM. We also implemented the AFR controller in the video streaming application by modifying ExoPlayer \cite{GithubExoPlayer}, which is an application-level media player for Android. The AFR controller collects the battery status of the mobile device and adds the battery level parameter when
transmitting a token to the video server for authentication.

NeuSaver utilizes the H.264/AVC video codec \cite{H264Overview}, which is widely used. The OpenH264 source code \cite{OpenH264, openH264Github} was modified to obtain the similarity intensities of the video chunks. The Y values of the macroblocks are extracted when the H.264/AVC encoder performs the video compression process. These extracted Y values are fed to the RL model of the AFR server as the similarity scores of the video chunks. For ease of implementation, we designed the video server to cover the CDN role that holds video content.
\noindent \textbf{NN architecture.} The RL model of the AFR server was implemented using TensorFlow \cite{abadi2016tensorflow}. NeuSaver fed the average Y values of nearby chunks, raw Y values, and size of the next chunk to 1D-CNN with 128 filters of size 4 with stride 1. These layers were then aggregated into three hidden layers with the same filters and stride. The original frame rate of the video and the previous decision of the frame rate were fed into a fully connected layer with 128 neurons. As the final output layer, the actor network applied the softmax function to this NN structure to generate the policy and the critic network applied a linear function to help the actor network to update the policy parameter. When training the RL model, the discount factor \(\gamma\), which determines how much the current choice will affect the future choice, was set to 0.99. The learning rates \(\alpha\) and \(\alpha'\) were both set to \(10^{-3}\). Lastly, the entropy factor \(\beta\), which determines the intensity of exploration, was configured to decay from 1 to 0.3 over \(10^{3}\) iterations. While tuning in other ways may be useful, the A3C method is known to be robust and stable to various settings \cite{mnih2016asynchronous}. Therefore, the parameters used in the existing A3C method were applied without meticulous tuning.

\noindent \textbf{AFR processor.}
After the streaming server generates several videos with different resolutions to provide the DASH service, the AFR processor processes these videos at various frame rate levels. To support smooth quality transitions with DASH, the AFR processor splits input videos based on instantaneous decoding refresh (IDR) frames to generate video chunks and then adjusts the frame rate within the video chunks that can be independently encoded. The video chunk length was set to 2 seconds in our experiment, as discussed in Section~\ref{sec:discussion}. The GOP of the video chunk consists of independently encoded I-frame and P and B-frames that are encoded intertwined with other frames. When processing the frame rate of the video chunk, only P or B-frames, except I-frames, were modified to make the chunking criteria of the video the same as before. In addition, the position of the key frame, which was an IDR frame, and GOP were adjusted in proportion to the target frame rate to be processed so that no noticeable jumps or stalls would occur during quality switching. These processes were performed through
FFmpeg \cite{FFmpeg} and were accomplished by individually re-encoding each video chunk.
Therefore, videos produced by NeuSaver could be smoothly switched with existing DASH-generated videos during streaming.



\section{Evaluation} \label{sec:evaluation}

We evaluated the performance of NeuSaver by examining the following questions: 1) How much can the average frame rate of processed videos be reduced? 2) How well can the quality of processed videos be maintained? 3) By how much can the energy consumption of mobile devices be reduced? 4) How much is the user experience affected? 5) How much system overhead is caused by NeuSaver? 

\subsection{Methodology} \label{subsec:Methodology}
\noindent \textbf{Video dataset.}
NeuSaver utilizes YouTube's video recommendation algorithm \cite{covington2016deep} to collect similar types of videos for training the RL model.
We trained the RL model with 571 videos from 4 categories: \textbf{News}, \textbf{Music Video}, \textbf{Game}, and \textbf{Cartoon}. For each category, we randomly selected 80\% of the video samples as a training set and used the remaining 20\% as a test set.

\begin{table}
\begin{center}
     \resizebox{1\columnwidth}{!}{$
    \begin{tabular}{|| c | c | c | c || c || c ||}
        \hline
        \thead{QoE\\Name} & \thead{\(\mu_{1}\)} & \thead{\(\mu_{2}\)} & \thead{\(\mu_{3}\)} &
        \thead{\(B(C^{r}_{k}\))} & \thead{\(P(C^{r}_{k}\))}\\ [0.5ex]
        \hline\hline
        \thead{\(QoE_{Q}\)} & \thead{7} & \thead{2.5} & \thead{14} &
        \thead{\(
    \begin{cases}
        5, & \text{if } C^{r}_{k} \geq 98 \\
        0, & otherwise
    \end{cases}\)} & \thead{\(
    \begin{cases}
        0, & \text{if } C^{r}_{k} \geq 90 \\
        15, & otherwise
    \end{cases}\)} \\
        \hline
        \thead{\(QoE_{B}\)} & \thead{4} & \thead{2} & \thead{17} &
        \thead{\(
    \begin{cases}
        2, & \text{if } C^{r}_{k} \geq 98 \\
        0, & otherwise
    \end{cases}\)} & \thead{\(
    \begin{cases}
        0, & \text{if } C^{r}_{k} \geq 85 \\
        15, & otherwise
    \end{cases}\)} \\
        \hline
    \end{tabular}
     $}
\caption{Parameter settings of Equation~\ref{eq:reward_function} for QoE metrics with different targets.}
\label{tab:QoE_setting}
\end{center}
\end{table}

\noindent \textbf{Reward function settings (\(QoE_Q\), \(QoE_B\)).}
Based on the results of multiple experiments, we propose \(QoE_Q\) and \(QoE_B\) and set the parameters of the reward function (Equation~\ref{eq:reward_function}), as shown in Table~\ref{tab:QoE_setting}. \(QoE_Q\) attempts to maintain the high video quality as much as possible by only reducing the frame rates of almost stationary video chunks. On the other hand, \(QoE_B\) scales down the frame rate of the video chunks more aggressively to reduce the energy consumption on mobile devices. We set it to operate as \(QoE_Q\) by default but set it to change to \(QoE_B\) when the sleep mode event occurs due to a low battery condition of the mobile device. This can be tuned differently depending on the user's purpose.

The VMAF metric is known to correlate with human perception better than other metrics such as VQM-VFD \cite{wolf2011video} and SSIM \cite{VMAF}. The VMAF methodology involves using the double stimulus impairment scale (DSIS) method to score the impairments in the processed video perceived by each observer, and the DSIS scores from all of the observers are combined to generate a differential mean opinion score and then normalized to a percentage in the range from 0 to 100. To minimize the user experience degradation, we set the criterion for the maximum penalty of \(QoE_Q\) to 90\%, which is the median score corresponding to an "imperceptible" DSIS score (5 points). Meanwhile, the criterion for the maximum penalty of \(QoE_B\) was set to 85\%, which is slightly below the penalty of \(QoE_Q\). Various additional QoEs can be defined depending on the purpose or environment in which NeuSaver is built, such as IoT infrastructure.


\subsection{Perceived Quality vs. Frame Rate}

\begin{table*}
\centering
     \resizebox{1\textwidth}{!}{$
\begin{tabular}{lccccccccc}
\toprule
\multirow{2}{*}{Category} & \multicolumn{2}{c}{\begin{tabular}[c]{@{}c@{}}\makecell{\(QoE_Q\)}\end{tabular}} &  \multicolumn{2}{c}{\begin{tabular}[c]{@{}c@{}}  \makecell{\(QoE_B\)}\end{tabular}} &
\multicolumn{2}{c}{\begin{tabular}[c]{@{}c@{}}\makecell{\(EVSO_B\) \cite{park2019evso}}\end{tabular}} &
\multicolumn{1}{c}{\begin{tabular}[c]{@{}c@{}} \makecell{Naive-60\%}\end{tabular}} &
\multicolumn{1}{c}{\begin{tabular}[c]{@{}c@{}} \makecell{Naive-40\%}\end{tabular}} &
\multirow{2}{*}{Type}
\\
\cmidrule(lr){2-3} \cmidrule(lr){4-5} \cmidrule(lr){6-7} \cmidrule(lr){8-8}
\cmidrule(lr){9-9}
& FPS (\%) & VMAF (\%) & FPS (\%) & VMAF (\%) & FPS (\%) & VMAF (\%) & VMAF (\%) & VMAF (\%) \\
\toprule
\textbf{News} & 51.96 & 98.61 & 43.50 & 97.52 & 63.07 & 98.99  & 98.35 & 96.88 & \textbf{A}\\ \toprule
\textbf{Game} & 86.17 & 98.68 & 76.92 & 95.60 & 78.39 & 95.20 & 86.95 & 78.32 & \textbf{B}\\ \toprule
\textbf{Music Video} & 73.18 & 96.71 & 66.24 & 94.02 & 77.37 & 94.77 & 90.03 & 82.39 & \multirow{2}{*}{\textbf{C}}\\
\textbf{Cartoon} & 76.95 & 96.71 & 69.25 & 93.07 & 73.81 & 93.01 & 84.66 & 75.17\\
 \bottomrule
\end{tabular}
     $}
\caption{Comparison of VMAF scores and FPS reduction when using \(QoE_Q\), \(QoE_B\), EVSO, Naive-60\%, and Naive-40\%.}
\label{tab:fps_vmaf_comparison}
\end{table*}

Table~\ref{tab:fps_vmaf_comparison} shows the reduced frame rate and VMAF score for test videos of categories processed using \(QoE_Q\), \(QoE_B\), \(EVSO_B\), and Naive. The videos were classified into three types according to the degree of movement: (A) static, (B) dynamic, and (C) hybrid.
Static videos consist of nearly identical scenes, such as those in a news interview. Dynamic videos consist of scenes with high intensity, such as war games. Hybrid videos contain both static and dynamic characteristics. For comparison, we created the state-of-the-art model \cite{park2019evso} called \(EVSO_B\) that determines an appropriate frame rate through the Y difference between the macroblocks. For performance comparison with \(QoE_B\), \(EVSO_B\) is parameterized so that its VMAF values are similar to the values maintained by \(QoE_B\). We also created the naive cases called Naive-60\% and Naive-40\%, in which the frame rates of all of the video chunks were simply reduced to 60\% and 40\%, respectively.

\(QoE_Q\), which prioritizes user experience, maintained the VMAF above 97\% on average while reducing the frame rate to about 72\% on average. On the other hand, \(QoE_B\), which focuses on reducing the battery consumption, decreased the frame rate to about 63\% on average, while keeping the VMAF scores above 93\%. The most important point here is that both QoEs adjusted the frame rate according to the motion characteristics of the video. Both QoEs decreased the frame rate of \textbf{News} category by 27\% more on average compared to the other categories. In contrast, the frame rates of \textbf{Game} category were conservatively reduced by 18\% less on average than the others. On the other hand, naively reducing the frame rate without taking into account the video characteristics can seriously degrade the user experience, especially for dynamic and hybrid videos. For example, Naive-40\% caused VMAF scores in \textbf{Game} and \textbf{Cartoon} categories to be less than 80\%.

Similar to NeuSaver, \(EVSO_B\) flexibly adjusted the frame rate according to the characteristics of the video, so there is no situation where the VMAF score drops below 90\%. However, \(QoE_B\) outperformed \(EVSO_B\) in most situations. The average VMAF score of \(EVSO_B\) and \(QoE_B\) was 95.88\% and 95.05\%, respectively, which are similar values, while \(QoE_B\) reduced the frame rate by about 9\% on average compared to \(EVSO_B\).

The most reason for the performance gap between EVSO and NeuSaver is the difference in the amount of information between EVSO and NeuSaver. To determine the frame rate of a video chunk, EVSO is based only on a single value of the sum of the Y difference values of internal frames, while NeuSaver utilizes the trained RL model based on various states such as raw Y difference values and video chunk sizes. In particular, we found that EVSO lowered the frame rate because the sum of Y difference values for scenes, such as when the camera is zoomed in or when small objects are actively moving, is relatively small. However, lowering the frame rate in these scenes severely dropped the VMAF score. On the other hand, NeuSaver maintains a relatively high frame rate in these scenes because it considers various characteristics using the RL model.


\begin{figure}[!t]
        \centering
    \includegraphics[width=1\columnwidth]{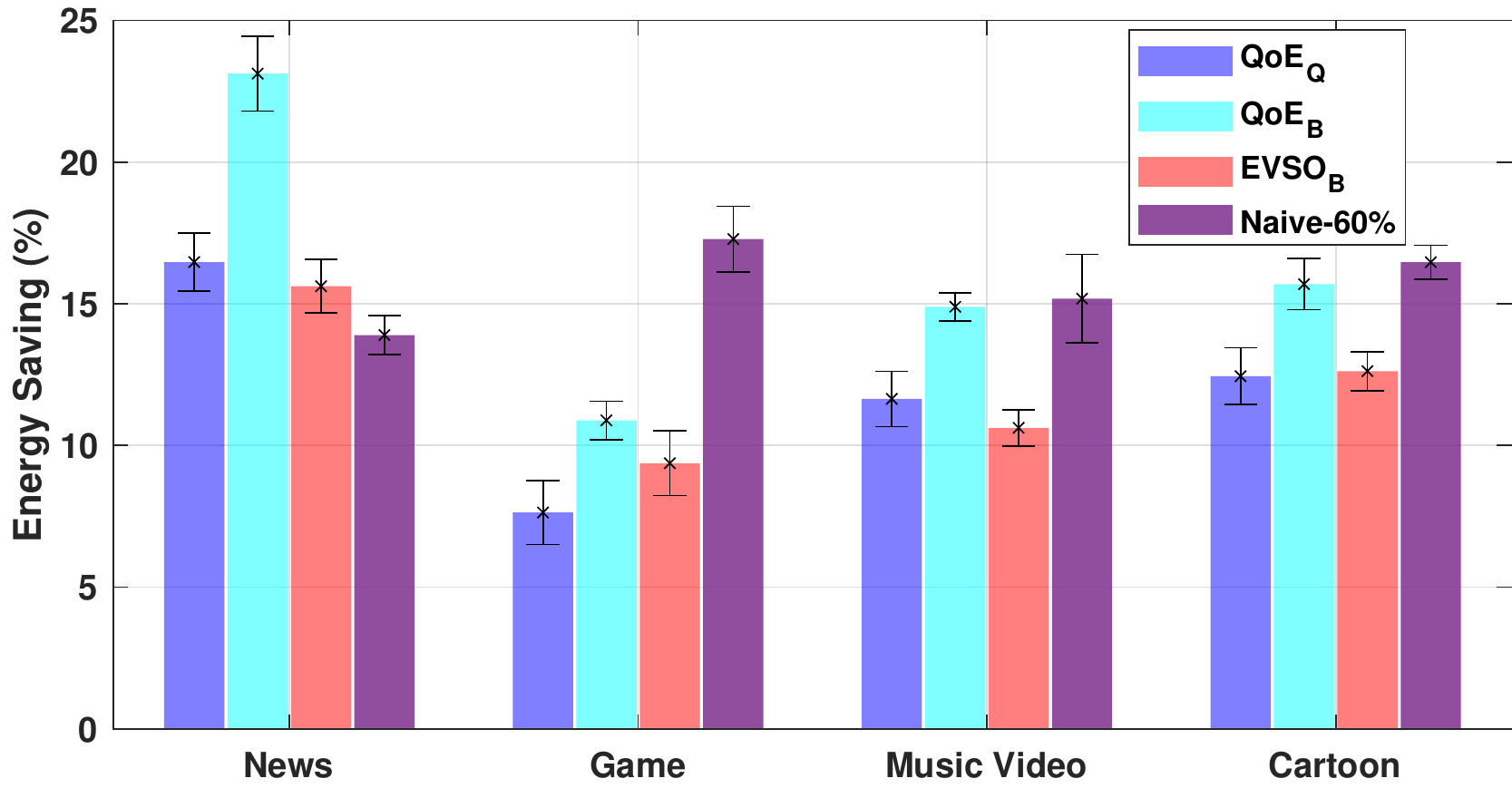}

    \caption{Reduction in energy usage while streaming each video.}
    \label{fig:energy_saving}
\end{figure}

\subsection{Energy Saving}
We used a high voltage power monitor produced by Monsoon \cite{Monsoon} to measure the energy consumption accurately when streaming videos on a mobile device. We chose an LG Nexus 5 smartphone as a mobile device. The screen brightness was set to 30\%, airplane mode was enabled to eliminate the effects of external variables, other applications were shut down except for the video streaming application, and Wi-Fi mode was set on to stream video from the server. We conducted all experiments while maintaining a high network bandwidth of about 160 Mbps. Note that we separated the decision process of the AFR Controller and the ABR Controller. Therefore, in our experiments, the ABR Controller always selects the best resolution for all video chunks. The mobile device was cooled down before each experiment to prevent thermal throttling due to the central processing unit overheating.

We randomly selected one video from each category and evaluated the power consumption of four videos with four different settings: Baseline, \(QoE_Q\), \(QoE_B\), \(EVSO_B\), and Naive-60\%. The baseline was the raw video and acted as a control. \(EVSO_B\) and Naive-60\% were also included for performance comparison. The energy consumption was measured four times and then averaged for accuracy.

Figure~\ref{fig:energy_saving} shows the energy reduction obtained by using \(QoE_Q\), \(QoE_B\), \(EVSO_B\), and Naive-60\%, compared to Baseline. Since NeuSaver adjusts the frame rate based on the degree of movement, the amount of energy consumption reduction achieved by using \(QoE_Q\) and \(QoE_B\) differed significantly for each category, unlike when Naive-60\% was used. \(QoE_Q\) reduced the energy consumption by less than 8\% when applied to \textbf{Game} category (dynamic), while achieving more than 16\% energy savings in \textbf{News} category (static). \(QoE_B\), which prioritizes battery saving, reduced the energy consumption by the same amount as or even more than Naive-60\%, except for \textbf{Game} category. \(QoE_B\) also achieved more energy savings in all categories compared to \(EVSO_B\) that maintains similar video quality, which is about 4.1\% difference on average. \(QoE_Q\) and \(QoE_B\) reduced energy consumption by 12.04\%, and 16.14\% on average, respectively. In particular, in the \textbf{News} category, \(QoE_B\) reduced energy consumption by 23.12 \%, saving nearly 8\% more energy than \(EVSO_B\).

\begin{figure*}[!t]
\begin{subfigure} {0.33\textwidth}
    \includegraphics[width=1\textwidth]{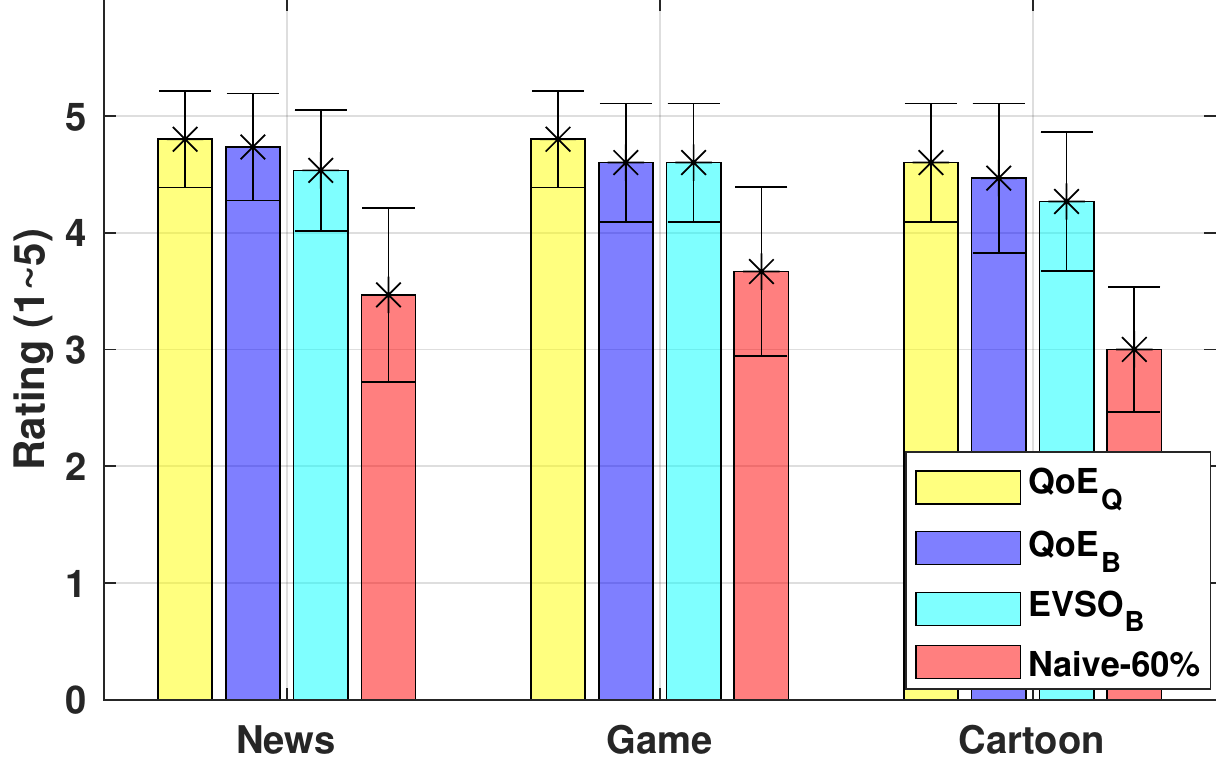}
    \caption{DSIS quality ratings.}
    \label{fig:DSIS}
\end{subfigure}
\begin{subfigure} {0.33\textwidth}
    \includegraphics[width=1\textwidth]{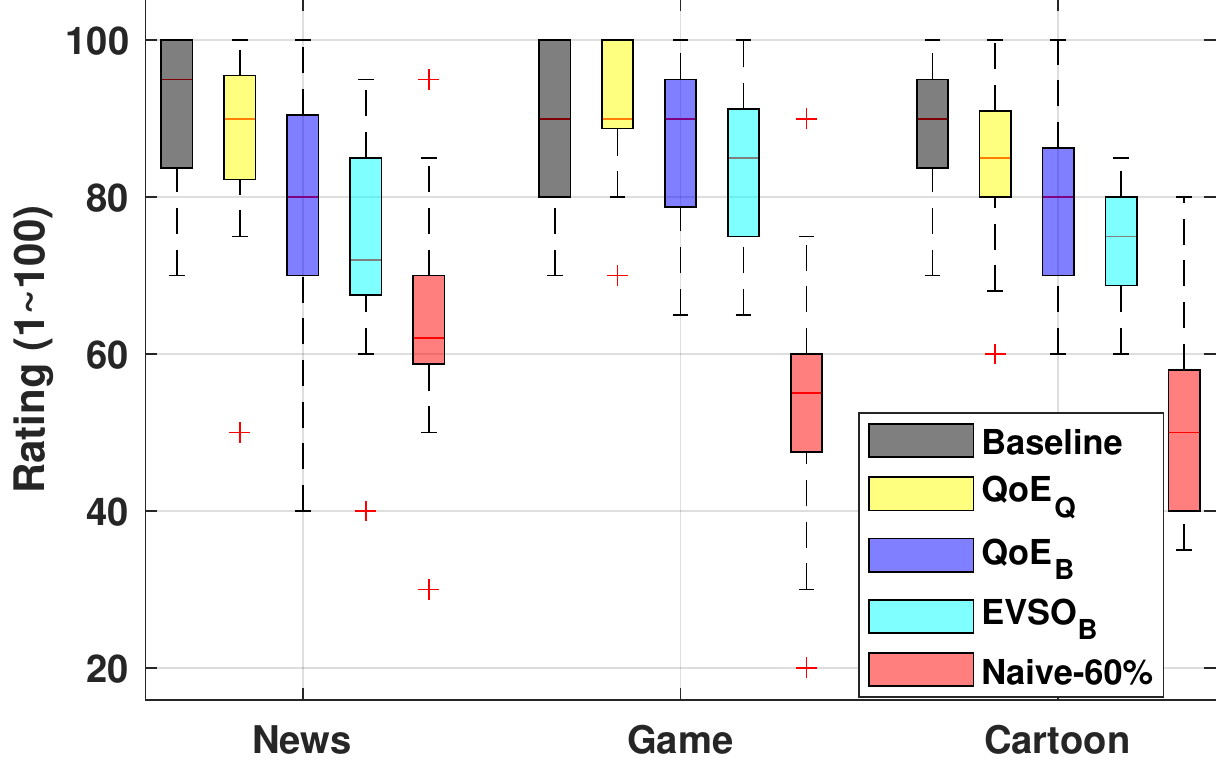}
    \caption{DSCQS quality ratings.}
    \label{fig:DSCQS_rating}
\end{subfigure}
\begin{subfigure} {0.33\textwidth}
    \includegraphics[width=1\textwidth]{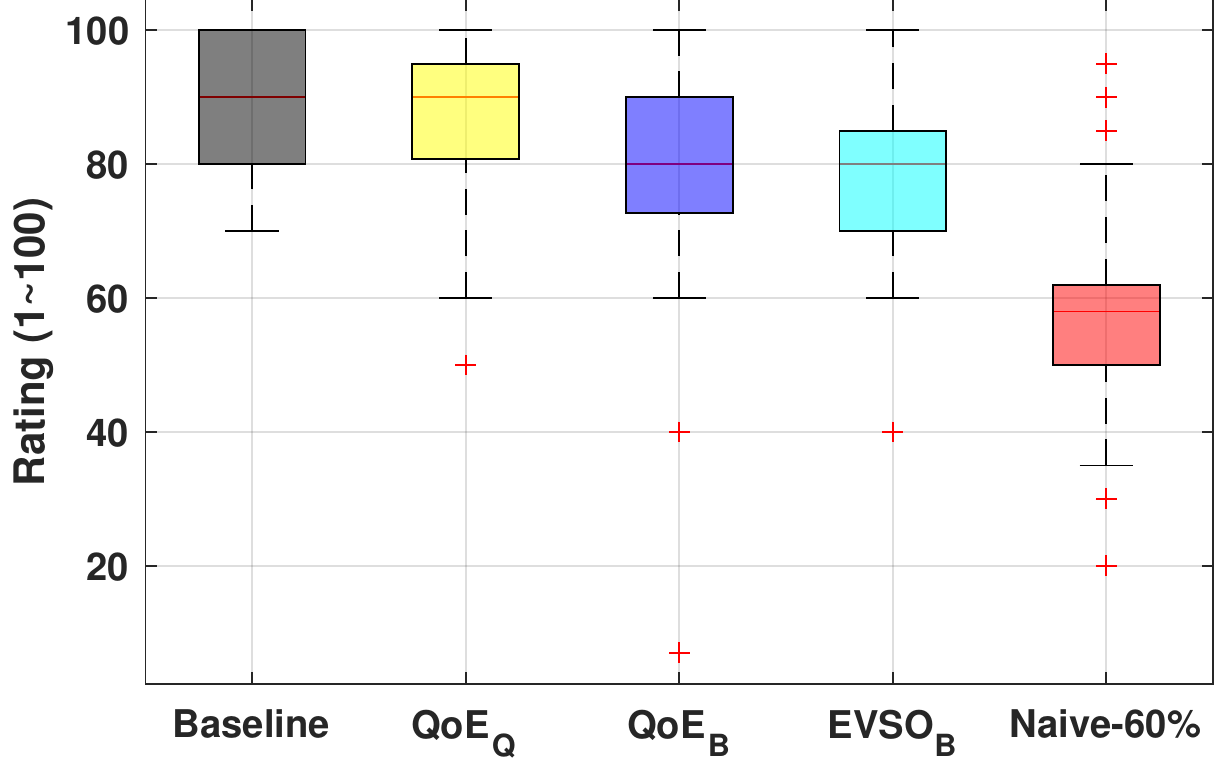}
    \caption{DSCQS quality rating distributions.}
    \label{fig:DSCQS_distribution}
\end{subfigure}
\caption{DSIS and DSCQS results from user study.}
\label{fig:user_study}
\end{figure*}

\subsection{User Study}
We recruited 16 participants (8 males) to analyze the effects of video processing using NeuSaver on the user experience. We employed the DSIS and the double stimulus continuous quality scale (DSCQS), which are commonly utilized methods of subjective quality assessments \cite{bt2002methodology, kim2016content, RAVEN}. The participants were instructed to watch three categories of videos: \textbf{News} (static), \textbf{Game} (dynamic), and \textbf{Cartoon} (hybrid). We instructed the participants to watch 1 minute of each video for accurate comparison.

\noindent \textbf{DSIS.}
In the DSIS method, the participants viewed the source and processed video sequentially, then evaluated the degree of impairment of the processed video. The DSIS rating index is a five-point impairment score of two stimuli ranging from very annoying (1 point) to imperceptible (5 points). DSIS provides participants with information about whether the video is the original or processed video.

We measured how much NeuSaver impaired the user experience. After we informed the participants which video was the processed one, the participants watched the original and processed videos sequentially.

Figure~\ref{fig:user_study}(a) shows the impairment ratings of three videos through the DSIS method. Most of the participants assigned scores of "imperceptible" (5 points) or "perceptible, but not annoying" (4 points) to videos processed with \(QoE_Q\) and \(QoE_B\) and \(EVSO_B\) compared to the original video. Thus, most of the participants perceived little or no quality degradation in the processed videos when compared to the original one. In addition, the overall high score for all types of videos indicates that NeuSaver properly adjusted the frame rates of the video chunks to match the motion intensity.

\noindent \textbf{DSCQS.}
For the DSCQS measurements, the participants viewed the source and the processed videos in random order to assess the difference in perceived visual quality. The DSCQS index is measured on a scale from 0 to 100. In contrast to DSIS, DSCQS does not involve informing the participants of the information in each video.

The participants watched the three types of videos as in the DSIS method. For each type, the participants randomly viewed each video processed using five settings: \(QoE_Q\), \(QoE_B\), \(EVSO_B\), Naive-60\%, and Baseline. The participants assessed the quality of each video after viewing all of the videos. We allowed the participants to watch the videos again if desired.

Figures~\ref{fig:user_study}(b) and (c) show the DSCQS quality ratings and the corresponding distribution of three videos processed using four settings. Most of the participants assigned similar scores to the original video (i.e., Baseline) and the video processed using \(QoE_Q\). \(QoE_B\)-processed videos were assigned a relatively low score than \(QoE_Q\) but had a high-quality rating with a difference of 10 points on average compared to Baseline. In addition, \(QoE_B\) had a quality rating of 3 points higher on average than \(EVSO_B\). On the other hand, most of the participants were able to differentiate \(QoE_B\) and Naive-60\% easily, which lowered the video's frame rate to almost the same level.
In addition, Figure~\ref{fig:user_study}(b) shows that the rating difference between Naive-60\% and \(QoE_B\) is greater in dynamic and hybrid videos than in static videos. The average rating gap between \(QoE_B\) and Naive-60\% was over 24 points. This indicates that NeuSaver not only reduces the power consumption as much as naively reducing the frame rate to 60\% but also preserves a high user experience.

\subsection{System Overhead}
All of the overhead experiments were repeated five times and the results were averaged for accuracy. The computer specifications were the same as those specified in Section~\ref{sec:implementation}.

\noindent \textbf{Perceptual similarity extraction.}
We measured how much additional time was spent on luminance extraction during the video encoding process based on a 2-minute video. The H.264/AVC encoder with and without the extraction process took about 22.02 seconds and 21.85 seconds on average, respectively. Therefore, the time overhead for similarity extraction was only minimal, at only  0.77\% on average.

\noindent \textbf{Training time of the RL model.}
The RL model consisting of 3 hidden layers and 128 filters requires more than 85,000 iterations to converge where each iteration took 510 ms. Therefore, the RL model total training time was approximately 12 hours when 16 agents were used in parallel. Since the RL model can be pre-trained, training overhead only occurred offline on the server-side, which supports NeuSaver. In addition, little effort is required to train the model after the RL model has stabilized.


\noindent \textbf{Communication between AFR modules.}
When a client attempts to stream a video, the AFR controller sends the current battery status of the mobile device to the AFR server to obtain the set of frame rates from the RL model. This additional communication overhead can be offset by forwarding this request when the client requests the video server to obtain information about the available bit rates and the location of the CDN hosting the video.

\noindent \textbf{Video processing of AFR processor.}
After the video server processes the original video with various bit rates for DASH functionality, the AFR processor additionally generates videos at various frame rates for the video with each resolution. We set five levels of frame rates in our experiments. Thus, a video server that supports NeuSaver incurs five times more processing and storage overhead than one that does not. However, the video processing is handled internally by the video server and can be done offline as preliminary work before streaming the video to the user. In addition, the processing and storage overhead can be significantly lessened if the AFR processing is operated based on some selective strategy, such as processing some popular videos. Therefore, the overhead caused by AFR processing can be tolerated by both the user and the video server. Note that NeuSaver is primarily intended to reduce battery consumption on mobile devices, so it is not mandatory to process every streamed video using NeuSaver.


\section{Discussion} \label{sec:discussion}
This section discusses several issues encountered when designing NeuSaver.

\subsection{Impact of AFR controller on ABR controller} \label{subsec:relationship}

\begin{figure}[!t]
      \centering
    \begin{subfigure} {1\columnwidth}
        \includegraphics[width=1\textwidth]{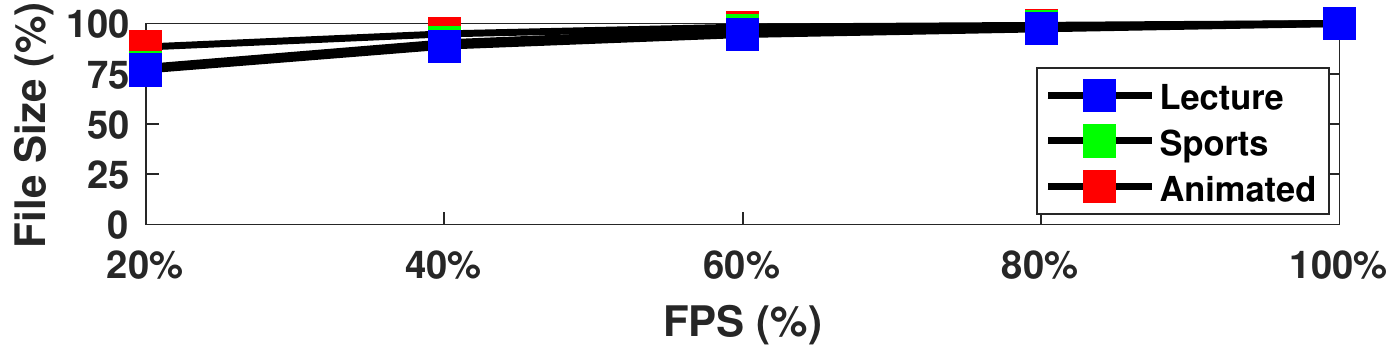}
        \caption{Correlation between FPS and file size.}
        \label{fig:fps_size}
    \end{subfigure}
    \begin{subfigure} {1\columnwidth}
        \includegraphics[width=1\textwidth]{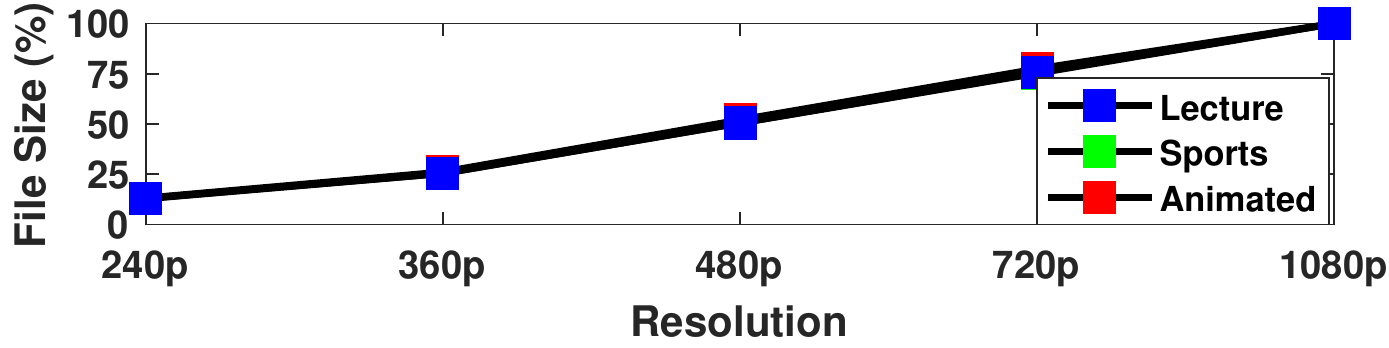}
        \caption{Correlation between resolution and file size.}
        \label{fig:resolution_size}
    \end{subfigure}

    \caption{Variation of file size according to video resolution and FPS change. The original FPS and resolution of the three videos are 30 and 1080p, respectively.}
    \label{fig:size_comparison}
\end{figure}

We designed NeuSaver so that the ABR controller and AFR controller are oblivious to each other. This separation allows the AFR controller to be mounted on the system without modifying the existing code corresponding to the ABR controller or the video manifest such as media presentation description (MPD). However, it may be optimal from a performance-wise point of view to consider the decisions of the AFR controller and the ABR controller harmoniously. For example, when the ABR controller requests a lower-resolution chunk because the available bandwidth is limited, if the AFR controller also requests a lower-FPS chunk because the video chunk has slow motion at the same time, the bandwidth would be underutilized. In this case, optimally, a higher resolution may be requested as much as the bandwidth saved by the AFR controller. However, bandwidth inefficiency caused by independent decisions between the ABR controller and AFR controller is insignificant.
Figure~\ref{fig:size_comparison}(a)
and (b) show how video resolution and FPS affect network bandwidth as it decreases. As the video resolution decreases, the video size decreases linearly at a similar rate. At the minimum resolution, it is reduced to about 12\% of the original video size, while at the minimum FPS the video size is about 81\% of the original video size on average. In other words, reducing the resolution by one step affects the network bandwidth more than reducing the FPS to the minimum. Therefore, we separated the two controllers in terms of efficiency of system integration rather than some performance inefficiency, but we can consider a method such as multi-task learning \cite{lu2017fully, caruana1997multitask} that trains separate tasks together to optimize the performance of both controllers in future work.



\renewcommand{\arraystretch}{0.2}
\begin{table}[h]
    \begin{subtable}[h]{0.45\textwidth}
    \centering
    \begin{tabular}{|| c | c ||}
        \hline
        \thead{Hidden units and filters} & \thead{Average \(QoE_Q\)} \\ [0.5ex]
        \hline\hline
        \thead{4} & \thead{-1.427}\\
        \hline
        \thead{16} & \thead{-1.346}\\
        \hline
        \thead{32} & \thead{-1.339}\\
        \hline
        \thead{64} & \thead{-1.317}\\
        \hline
        \thead{\textbf{128}} & \thead{\textbf{-1.303}}\\
        \hline
        \thead{256} & \thead{-1.302}\\
        \hline
    \end{tabular}
    \caption{Average \(QoE_Q\) with different numbers of hidden units. One hidden layer was used.}
    \label{tab:NN_parameter_settings_units}
    \end{subtable}
    \hfill
    \begin{subtable}[h]{0.45\textwidth}
    \centering
    \begin{tabular}{|| c | c ||}
        \hline
        \thead{Hidden layers} & \thead{Average \(QoE_Q\)} \\ [0.5ex]
        \hline\hline
        \thead{1} & \thead{-1.303}\\
        \hline
        \thead{2} & \thead{-1.238}\\
        \hline
        \thead{\textbf{3}} & \thead{\textbf{-1.187}}\\
        \hline
        \thead{4} & \thead{-1.177}\\
        \hline
    \end{tabular}
        \caption{Average \(QoE_Q\) with different numbers of hidden layers and 128 hidden filters.}
        \label{tab:NN_parameter_settings_layers}
    \end{subtable}
\caption{Performance comparison of \(QoE_Q\) with various NN parameter settings. \(QoE_Q\) was calculated by averaging four experiments in each case.}
\label{tab:NN_parameter_settings}
\end{table}

\subsection{NN Parameter Settings} \label{subsec:Parameter_settings}
We experimented with various combinations of NN parameters to analyze how the NN parameters of the RL model affect the performance of NeuSaver and to identify the optimal settings. Table~\ref{tab:NN_parameter_settings}(a) shows the average \(QoE_Q\) according to the number of hidden units with one hidden layer. The average \(QoE_Q\) gradually improves as the number of filters increases. However, the average \(QoE_Q\) stagnated when 128 filters were used and almost the same performance was obtained as when 256 filters were used, so we set the default filter number of NeuSaver to 128. Table~\ref{tab:NN_parameter_settings}(b) shows the average \(QoE_Q\) based on the number of hidden layers with 128 filters. We performed training until the NN model converged under each condition. Approximately, 20,000 additional iterations were required for each hidden layer. As can be seen from Table~\ref{tab:NN_parameter_settings}(b), the average \(QoE_Q\) significantly improves as the number of hidden layers increases. However, much more training time and training data are required as the number of hidden layers increases. Therefore, the number of hidden layers should be determined according to the complexity of the problem to be solved. We set the default number of hidden layers to three in NeuSaver. The NN parameter settings can be finely tuned depending on the environment in which NeuSaver is used.

\subsection{Trade-off between encoding efficiency and flexibility} \label{subsec:Trade}
When a client requests a video from a video server, the RL model determines the optimal frame rate for the chunks of that video. The video chunk length should be determined by making a compromise between encoding efficiency and flexibility for stream adaptation to changes in motion intensity. Since the first frame of a video chunk must always be an IDR frame that contains the entire image, if the video chunks are too short, the encoding process may be inefficient because many IDR frames that are not necessary for the situation may be generated. Meanwhile, if the video chunks are too long, the features within one video chunk may be ambiguous, which could cause difficulty in determining the proper frame rate of the video chunk. To achieve excellent processing performance using MPEG-DASH, YouTube recommends a video chunk length of 1-5 seconds \cite{YouTubeRecommendChunkLength}. Considering all of these factors, we set the video chunk length to 2 seconds by balancing the motion characteristics of the video chunk and codec performance.






\section{Related Work} \label{sec:related}

This section reviews the existing works related to energy reduction on mobile devices while streaming videos and various approaches using RL models.

\subsection{Energy Saving when Streaming Videos} \label{subsec:Energy}
EVSO \cite{park2019evso} is a system that adaptively adjusts the frame rates of each part of the video according to the motion intensity. Using luminance values, EVSO splits the video into multiple chunks with similar degrees of motion intensity, then adjusts the frame rate of each chunk to the appropriate frame rate. This system can lower the frame rates of static video parts, significantly reducing the energy consumption with little effect on the user experience. However, EVSO did not take into account the GOP structure, IDR frame position, decoding time stamp (DTS), and presentation time stamp (PTS). If a video is arbitrarily segmented using the EVSO method, the processed video contains more unnecessary IDR frames than the original video and the order of DTS and PTS is messed up, which can cause inefficiencies in the encoding and decoding processes. In addition, since the processed videos are larger than the original ones due to the increased number of IDR frames, streaming these videos is inefficient for network bandwidth and can cause additional energy consumption by increasing the operating time of the wireless interface.

Lim et al. \cite{lim2016adaptive} proposed an adaptive frame rate control scheme using an H.264/AVC encoder for multimedia-sharing applications. This approach achieves energy savings for mobile devices while maintaining high video quality by skipping frames with low motion intensities during the encoding process of the server. In addition, this scheme is advantageous for the network bandwidth of the client since skipped frames are not transmitted to the mobile device. However, this approach is impractical for a video streaming service in which a large number of new video content is uploaded every day because it uses the SSIM method, which incurs a high computational overhead when calculating the similarity of frames. NeuSaver, on the other hand, is a specialized model for video streaming services and provides a policy to determine the appropriate frame rate through various video features, including the Y value of macroblocks.


Since wireless interfaces consume significant amounts of energy on mobile devices, considerable efforts have been made to optimize the operating times of wireless interfaces when streaming videos \cite{rao2011network, hu2015energy}. These schemes extend the idle time of the wireless interface by pre-downloading future content and storing it in the playback buffer. These approaches can interoperate with NeuSaver; however, if a user skips or turns off videos frequently while watching, pre-downloaded content becomes useless and wastes network bandwidth.

Kim et al. \cite{kim2016content} proposed using the content rate, i.e., the number of meaningful frames per second, to eliminate redundant frames by adjusting the refresh rate of the display hardware. This scheme reduces the refresh rate in areas with low content rates, reducing the overall power consumption while maintaining a satisfactory user experience. However, since this approach only adjusts the refresh rate on the user side, unnecessary frames that are not to be used are sent to the user, causing unnecessary energy consumption.


\subsection{RL Approaches} \label{subsec:Reinforcement}
Pensieve \cite{mao2017neural} is a system that uses the RL model to determine the bit rates of future video chunks. Since existing ABR algorithms use fixed control rules or models, it is difficult to provide optimal solutions in various network conditions. On the other hand, Pensieve automatically reinforces the model in response to the feedback from the environment without pre-programmed models or rules. Consequently, Pensieve outperformed existing ABR algorithms in various network conditions.

Similar to Pensieve, QARC \cite{huang2018qarc} used a deep RL model to adapt to various network conditions when determining the bit rate of the next video chunk. The difference from Pensieve is that the video quality metric is added to the reward function considering the user experience when determining the bit rate of the next video chunk. QARC outperformed the existing rate control algorithms in various network conditions and QoE goals.


\section{Conclusion and Future Work} \label{sec:conclusion}
We presented a system called NeuSaver that can optimize the energy consumption of mobile devices in video streaming services. NeuSaver reduced the overall frame rate of the video by applying an adaptive frame rate to the video chunk. To determine the proper frame rate for each video chunk, we used reinforcement learning (RL) to learn the optimal policy that can achieve QoE goals. In addition, we trained the RL model effectively and robustly by applying the asynchronous advantage actor-critic (A3C) algorithm. We fed the RL model with various inputs that can estimate the motion intensity of the video chunk. Our experiments showed that NeuSaver can reduce the energy consumption of mobile devices by up to 23.12\% while maintaining a high QoE.

In the future, we plan to expand and develop our RL model by applying several tactics for policy improvement. One potential approach would be to encourage agents to forage and explore more diverse situations to learn near-optimal policies.




\ifCLASSOPTIONcaptionsoff
  \newpage
\fi

\ifCLASSOPTIONcompsoc
  \section*{Acknowledgments}
\else
  \section*{Acknowledgment}
\fi
This paper is an extended version of the authors’ previous work published in the \textit{Proceedings of IEEE INFOCOM 2019~}\cite{park2019evso}. The authors would like to thank the anonymous reviewers and colleagues for their valuable comments.


%
%

\bibliographystyle{IEEEtran}
\bibliography{bibliography}

%
\newpage
\begin{IEEEbiography}[{\includegraphics[width=1in,height=1.25in,clip,keepaspectratio]{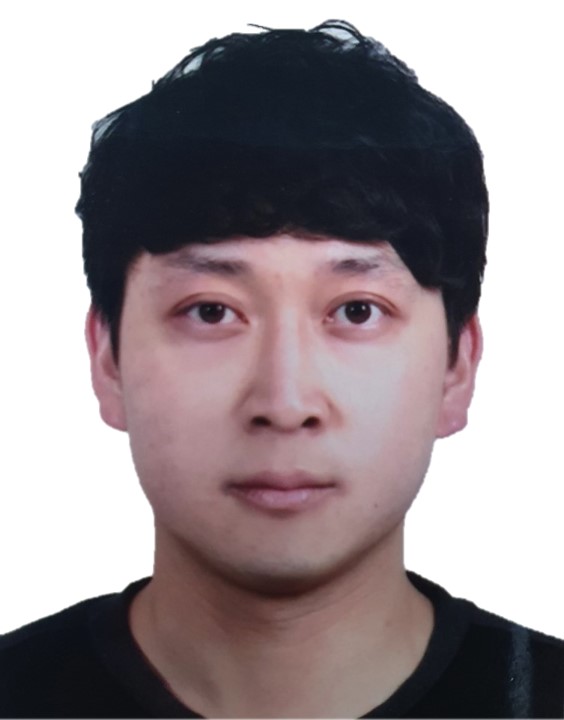}}]{Kyoungjun Park}
received his B.S. degree in computer science engineering from Chung-Ang University, Seoul, South Korea, in 2017, and M.S. degree in school of computing, Korea Advanced Institute of Science and Technology (KAIST), Daejeon, South Korea, in 2019. As alternative military service, he is currently working as a research engineer at TmaxSoft, Seongnam, South Korea. His research interests include mobile and ubiquitous systems, machine learning, reinforcement learning, and human-computer interaction.

\end{IEEEbiography}

\begin{IEEEbiography}[{\includegraphics[width=1in,height=1.25in,clip,keepaspectratio]{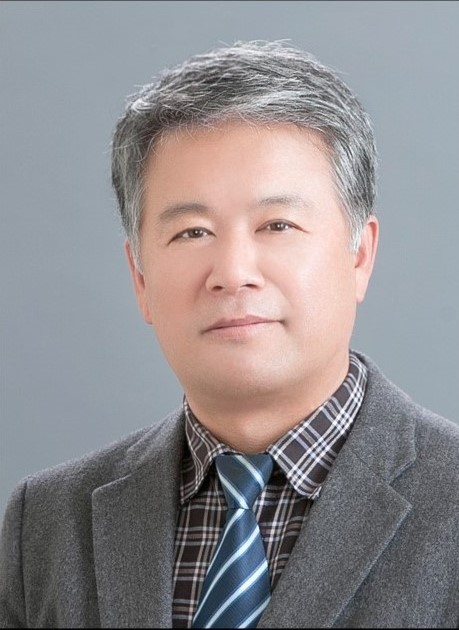}}]{Myungchul Kim}
received his B.A. in Electronics Engineering from Ajou University in 1982, M.S. in Computer Science from the Korea Advanced Institute of Science and Technology (KAIST) in 1984, and Ph.D. in Computer Science from the University of British Columbia, Vancouver, Canada, in 1993. Currently, he is with the faculty of the KAIST as Professor in the School of Computing. Before joining the university, he was a managing director in Korea Telecom Research and Development Group during 1984–1997 where he was in charge of research and development of protocol and QoS testing on ATM/B-ISDN, IN, PCS and Internet. He has also served as a member of Program Committees for numerous numbers of conferences and served as chair of the IWTCS'97 and the FORTE’01. He has published over 150 conference proceedings, book chapters, and journal articles in the areas of computer networks, wireless mobile networks, protocol engineering and network security. His research interests include Internet, protocol engineering, mobile computing, and information security.
\end{IEEEbiography}


\begin{IEEEbiography}[{\includegraphics[width=1in,height=1.25in,clip,keepaspectratio]{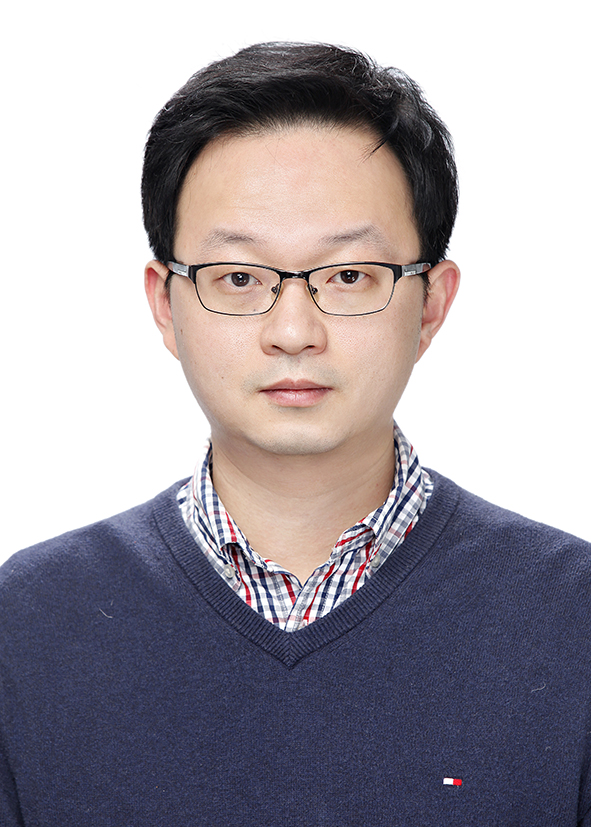}}]{Laihyuk Park}
received the B.S., M.S., and Ph. D. degrees in computer science and engineering from Chung-Ang University, Seoul, South Korea, in 2008, 2010, and 2017, respectively. From 2011 to 2016, he was a research engineer with Innowireless, Bun-Dang, South Korea. From 2018 to 2019, he is an assistant professor with Chung-Ang University, Seoul, South Korea. He is currently an assistant professor with the department of computer science and engineering, Seoul National University of Science and Technology (Seoultech), Seoul, South Korea. His research interests include the smart grid, the Internet of things, and next-generation networks.
\end{IEEEbiography}

\end{document}